\documentclass[conference]{IEEEtran}
\IEEEoverridecommandlockouts
\usepackage[T1]{fontenc}
\usepackage{graphicx}
\usepackage{makeidx} 
\usepackage{amsmath}
\usepackage[utf8]{inputenc}
\usepackage{amssymb}
\usepackage[english]{babel}
\usepackage{float}
\usepackage{enumitem}
\usepackage{tabularx}
\usepackage{booktabs}
\renewcommand{\baselinestretch}{1.15}
\usepackage{csquotes}
\usepackage[table]{xcolor}% http://ctan.org/pkg/xcolor
\usepackage{commath}
\usepackage{url}
\usepackage[colorlinks=false]{hyperref}
\usepackage{algorithmic}
\usepackage[ruled,vlined,linesnumbered]{algorithm2e}
\usepackage{comment}
\usepackage{mathtools}
\usepackage{placeins}
\usepackage{glossaries}
\usepackage{caption}
\usepackage{subcaption}
\usepackage{array}
\usepackage{colortbl} % For coloring cells
\usepackage{xcolor} % Define colors
\usepackage{adjustbox}
\usepackage{stfloats}
\newtheorem{definition}{Definition}

\usepackage{amsmath,amssymb,amsfonts}
\usepackage{algorithmic}
\usepackage{graphicx}
\usepackage{textcomp}
\usepackage{xcolor}
\def\BibTeX{{\rm B\kern-.05em{\sc i\kern-.025em b}\kern-.08em
    T\kern-.1667em\lower.7ex\hbox{E}\kern-.125emX}}
  
\begin{document}

\title{A Scalable and Near-Optimal Conformance Checking Approach for Long Traces}

\begin{comment}
\author{\IEEEauthorblockN{Eli Bogdanov}
\IEEEauthorblockA{\textit{dept. name of organization (of Aff.)} \\
\textit{name of organization (of Aff.)}\\
City, Country \\
email address or ORCID}
\and
\IEEEauthorblockN{Izack Cohen}
\IEEEauthorblockA{\textit{dept. name of organization (of Aff.)} \\
\textit{name of organization (of Aff.)}\\
City, Country \\
email address or ORCID}
\and
\IEEEauthorblockN{Avigdor Gal}
\IEEEauthorblockA{\textit{dept. name of organization (of Aff.)} \\
\textit{name of organization (of Aff.)}\\
City, Country \\
email address or ORCID}
}
\end{comment}

\author{\IEEEauthorblockN{Eli Bogdanov}
\IEEEauthorblockA{\textit{Data \& Decision Sciences} \\
\textit{Technion--Israel Institute of Technology}\\
Haifa 3200003, Israel \\
eli-bogdanov@campus.technion.ac.il\\
orcidID: 0000-0002-5084-7344}
\and
\IEEEauthorblockN{Izack Cohen}
\IEEEauthorblockA{\textit{Faculty of Engineering} \\
\textit{Bar-Ilan University}\\
Ramat Gan 5290002, Israel \\
izack.cohen@biu.ac.il\\
orcidID: 0000-0002-6775-3256}
\and
\IEEEauthorblockN{Avigdor Gal}
\IEEEauthorblockA{\textit{Data \& Decision Sciences} \\
\textit{Technion--Israel Institute of Technology}\\
Haifa 3200003, Israel \\
avigal@technion.ac.il\\
orcidID: 0000-0002-7028-661X}
}
\maketitle

\begin{abstract}

Long traces and large event logs that originate from sensors and prediction models are becoming more common in our data-rich world. In such circumstances, conformance checking, a key task in process mining, can become computationally infeasible due to the exponential complexity of finding an optimal alignment. 

This paper introduces a novel sliding window approach to address these scalability challenges while preserving the interpretability of alignment-based methods. By breaking down traces into manageable subtraces and iteratively aligning each with the process model, our method significantly reduces the search space.

The approach uses global information that captures structural properties of the trace and the process model to make informed alignment decisions, discarding unpromising alignments even if they are optimal for a local subtrace. This improves the overall accuracy of the results.

Experimental evaluations demonstrate that the proposed method consistently finds optimal alignments in most cases and highlight its scalability. This is further supported by a theoretical complexity analysis, which shows the reduced growth of the search space compared to other common conformance checking methods. 

This work provides a valuable contribution towards efficient conformance checking for large-scale process mining applications.

\end{abstract}


\section{Introduction}\label{sec:intro}
This work focuses on developing a scalable conformance checking approach that can handle lengthy traces containing hundreds or thousands of events, which traditional methods struggle to manage \cite{van2022process}. Technological advancements and widespread usage of sensors, the Internet of Things (IoT), and prediction models have resulted in an abundance of process data, thereby generating traces with thousands of events \cite{bogdanov2022conformance, bogdanov2023sktr}. 
Performing conformance checking on these traces has been identified as one of the foremost process mining challenges of our time \cite{beerepoot2023biggest}. Additionally, many industries, such as construction, aerospace, and infrastructure, involve large-scale processes that are realized as long traces \cite{cohen2024data}.

By breaking down a long trace into manageable subtraces and applying a sliding window technique that takes into account structural properties of the model and the trace, we present a scalable and efficient solution to the conformance checking problem with long traces. Each subtrace is aligned with the process model, leveraging the concept of iterative alignment. The process model retains its state from one subtrace alignment to the next, ensuring a continuous and coherent analysis. The proposed technique can be adjusted by introducing user-defined hyperparameters for controlling the balance between computational efficiency and alignment accuracy.

The suggested approach uses global information about the trace and model to make informed subtrace-level decisions. The idea is to take into account the subsequent implications of a local alignment by computing a lower bound on the marginal cost of a specific subtrace based on the number of activities remaining in the trace that cannot be executed from each state in the model. This cost is adjusted by the frequency of the nonreachable transitions within the remaining portion of the trace.

The suggested approach is scalable, as it limits search space growth relative to trace length. Performance measurements across multiple datasets show that it finds the optimal solution for over 96\% of traces, with an average deviation from the optimal cost of 0.66\%. Our tests demonstrate that this approach can handle traces with thousands of transitions—beyond the capability of most other methods—within reasonable running times.

The main contributions of this work are as follows:
\begin{enumerate}[label=\arabic*.]
    \item \textbf{Modeling and algorithm development:} We introduce a novel conformance checking approach that decomposes long traces while maintaining the process model intact. Theoretical characteristics are presented, showing how segmentation facilitates efficient handling of complex event logs without compromising the integrity of the overall process analysis. The algorithm accounts for various nuances to efficiently maintain and evaluate multiple alignment paths within the process graph.
    \item \textbf{Demonstration of scalability:} Complexity analyses of the developed algorithm demonstrate its superior scalability compared to other alternatives.
    \item \textbf{Empirical validation:} Experimental results on both known datasets and datasets with very long traces highlight the algorithm's performance. The experiments show reduced computational overhead and near-optimal alignment accuracy compared to alternative methods.
\end{enumerate}

The remainder of the paper is organized as follows. The next section presents definitions and the modelling framework. Section~\ref{sec:algorithm} formalizes the algorithm. Section~\ref{sec:complexity} presents a complexity analysis. Section~\ref{sec:empirical_evaluation} describes the experiments and their results followed by Section~\ref{sec:literature_review} that reviews key related studies. The last section concludes the paper and suggests avenues for future research.

\section{Modeling and Definitions}\label{sec:model}

%Long traces, sequences of events spanning hundreds or thousands of activities, pose a computational challenge to traditional conformance checking methods, which often struggle to efficiently process them, leading to analysis time and resource consumption bottlenecks~\cite{van2022process}. 

Our focus is on alignment-based conformance checking approaches \cite{adriansyah2014aligning} that can highlight how and where process executions diverge from the process model.

At the heart of such methods is the trace model, a chronologically ordered sequence of events, captured from log data. We rely on common definitions and notation of a trace model, synchronous product, cost function and optimal alignment (see, for example \cite{carmona2018conformance}) to define components of the suggested approach such as a \textit{subtrace} and \textit{partial optimal alignment}. %Definitions~\ref{def:trace_model}-\ref{def:opt_alignment} are based on \citeauthor{carmona2018conformance} \cite{carmona2018conformance}.

Typically, an optimal alignment is found by searching using $A^*$-based methods over the reachability graph of a synchronous product by progressing from its initial to final markings. At each explored marking, the algorithm sums the cost to reach that marking with a lower bound, calculated by a heuristic function, of the cost from the current marking to the final one. In other words, the algorithm maps lower bound costs and priorities the exploration of promising paths until finding the optimal alignment. 
%When applied, the algorithm evaluates the synchronous product of a trace and a process model, balancing the current alignments' accrued costs against the forecasted costs to secure the optimal alignment. This results in a preference for paths promising the lowest overall costs. 
%Consequently, the algorithm pinpoints an alignment 
%that not only faithfully reconstructs the actual event sequence but also 
%minimizes deviations from the process model. 
Despite $A^*$'s computational efficiency, finding an optimal alignment for long traces is computationally demanding. %This acknowledgment underscores the ongoing challenge in process mining to reconcile the demands of extensive trace analysis with the practicalities of computational resources and algorithmic limitations.
We bound the computational effort by partitioning the trace into {\em subtraces}, each of which is aligned with respect to the model. Similar decomposition and look-ahead ideas have been found efficient in problems that aim to find the shortest path under a limited computational budget (e.g., \cite{cohen2016discretization}). 

For the decomposition, we define the notions of a {\em subtrace model}, and {\em partial optimal alignment}.

%By dividing a long trace into subtraces, we enable the alignment of each subtrace with respect to the process model. Aligning the smaller subtraces with the model is significantly less computationally demanding compared to a complete alignment of the trace. %the fmodel adapts the structure of the original trace model to these smaller segments, reducing computational complexity by focusing on manageable sections.

\begin{sloppypar}
\begin{definition}[Subtrace Model] \label{def:subtrace_model}
Let $A \subseteq \mathcal{A}$ be a set of activities over the universe of all possible activities $\mathcal{A}$, and $\sigma \in A^*$ be a sequence from the set of all possible activity sequences. Given a trace model $TN =((P,T,F,\lambda),m_i,m_f)$, a subtrace model $STN =((P_s,T_s,F_s,\lambda_s),m_{is},m_{fs})$ is a system net for a subsequence $\sigma_s = \sigma(j:k)$ for some $0 \leq j < k \leq |\sigma|$, such that $P_s = \{p_j,...,p_k\}$ is the set of places, $T_s \subseteq \{t_{j+1},...,t_k\}$ is the set of transitions, $F_s = \{(p_i,t_{i+1}) | j \leq i < k\} \cup \{(t_i,p_i)| j+1 \leq i \leq k\}$ is the set of flows, and $m_{is} = [p_j]$ and $m_{fs}=[p_k]$, are the initial and final markings, respectively. It holds that $\lambda_s(t_i)=\sigma(i)$ for $j+1 \leq i \leq k$, where $\lambda_s$ is a labeling function that assigns an activity label to each transition in the subtrace, mirroring the labeling function $\lambda$ in the full trace model while restricting it to subsequence $\sigma_s$.
\end{definition}
\end{sloppypar}
%To complement this, we introduce the notion of partial optimal alignment (def. \ref{def:partial_opt_alignment}). 

We now use the notion of a subtrace to define a partial optimal alignment. Standard alignment-based conformance checking searches for an optimal alignment over the entire synchronous product, while a partial optimal alignment focuses on finding the best alignment for each subtrace taking into account the context of the entire process. This technique accommodates varying process states encountered across subtraces, starting with an initial state that merges the subtrace's beginning with an applicable state of the process model. The goal is to minimize deviation costs, culminating in a final state that combines the subtrace's end with any legitimate final state of the model. This approach streamlines conformance checking by sequentially addressing each subtrace. Here is a formal definition of a partial optimal alignment.

\begin{definition}[Partial Optimal Alignment] \label{def:partial_opt_alignment}
Let $A \subseteq \mathcal{A}$ be a set of activities, and $\sigma_s \in A^*$ represents a subtrace within a complete trace $\sigma$, with $STN =((P_s,T_s,F_s,\lambda_s),m_{is},m_{fs})$ as its corresponding subtrace model. Let $SN$ be a process model, and $SP_s$ the synchronous product of $SN$ and $STN$, reflecting the interaction between the process model and the subtrace model. Further, let $c : T \rightarrow \mathbb{R}^+ \cup \{0\}$ be a cost function. A partial optimal alignment $\gamma^{opt}_s \in L_{SP_s}$ is a sequence within the synchronous product that represents the lowest cost execution sequence starting from an initial marking $m_{init}$, which is the union of the initial marking $m_{is}$ in the $STN$ and a specific marking $m{'}$ of the $SN$, to a set of final markings. Each final marking is the union of the final marking $m_{fs}$ from $STN$ and any valid marking $m{''}$ within $SN$.
For all $\gamma_s \in L_{SP_s}$, it holds that $c(\gamma_s) \geq c(\gamma^{opt}_s)$, where $c(\gamma_s) = \sum_{1\leq i \leq  |\gamma_s|} \, c(\gamma_s(i))$. 
\end{definition}

\section{Algorithmic Design and Implementation}\label{sec:algorithm}
Building on the ideas in the previous section, we formalize the approach as Algorithm~\ref{alg:WindowBasedConformanceChecking} for conformance checking over long traces. 
We start by introducing the overall approach (Subsection~\ref{sec:approach}), followed by introducing the cost calculation (Subsection~\ref{sec:heuristic}) and the sliding window mechanism (Subsection~\ref{sec:sliding}). Algorithm~\ref{alg:WindowBasedConformanceChecking} presents the pseudocode to which references are made throughout.

\subsection{The Algorithmic Approach}\label{sec:approach}

A long trace is sliced into subtraces (see Definition \ref{def:subtrace_model}) and the algorithm progresses using a `window' that iteratively slides along the trace from one subtrace to its successor. The window length has to be sufficiently long to encompass relevant process activities yet not too long to ensure computational efficiency.

Decomposing the conformance checking problem into smaller subproblems is just one facet of our approach. To fully capitalize on this division and achieve an effective balance between computational efficiency and alignment accuracy, we couple the sliding window mechanism with an iterative alignment process that takes into account global information. The idea is that we could identify situations in which an alignment that is optimal for the local window may lead to high future costs, and in such a case, we could select a different alignment. %This dual strategy addresses the challenges of aligning long, complex traces, while also ensuring that the solution remains flexible and adaptable to the specific needs of the user.

\subsection{Embedding Local and Global Information}\label{sec:heuristic}

To enhance the alignment process, we leverage local and global information. The idea is to analyze the process model for mapping the reachable transitions from each marking. This knowledge is used during the alignment process to establish a lower bound on the additional cost, beyond the local window's conformance cost, associated with each potential alignment. This marginal cost reflects the minimal impact of a selected alignment on the overall conformance cost due to the structure of the process model and the trace beyond the local window while considering the markings resulting from specific alignment choices. 
For example, it is easy to calculate the number of activities remaining in the trace that cannot be executed from each state in the model which can be translated into nonsynchronous moves and thus to a marginal future cost. This cost can be weighted by the frequency of these nonreachable transitions within the remaining portion of the trace.

By considering the marginal cost from alignments that lead to states where synchronous moves are impossible for frequent transitions, the approach enhances the accuracy of the search for an optimal alignment and increases efficiency by eliminating the less promising alignments.

Let us illustrate the idea using a toy example of the process model in Figure \ref{fig:processmodelrunningexample} and the trace model in Figure \ref{fig:tracemodelrunningexample}. Assume that the process model is in marking $[p_{3}]$, i.e., a single token appears in place $p_{3}$, and the marking within the trace model is $[p'_{5}]$. Thus, the marking corresponding to their synchronous product is $[p_{3}, p'_{5}]$.

At this stage a first choice would be to perform a synchronous move $(E,E)$ with a zero cost that would lead to marking $[p_{4}, p'_{6}]$. In such a case, the algorithm would return 3 as a lower bound for the marginal cost, since from $p_4$ there are no reachable transitions but there are three transitions that must be executed in the trace corresponding to labels $C, C, E$ which would be performed as nonsynchronous moves. A second choice would be to execute model transition $\tau$ resulting in marking $[p_{2}, p'_{5}]$. The cost of this move is 0 and the lower bound marginal cost would also be 0 since the remaining transitions within the trace $(E, C, C, E)$ are reachable from this marking. Indeed, as we will demonstrate in the following subsections, the second choice would be preferable over the first one. 

\begin{algorithm}[ht]
    \small
    \SetInd{0.1em}{0.3em}
    \SetKwInOut{Input}{Input}
    \SetKwInOut{Output}{Output}
    \SetKwInOut{Hyperparameters}{Hyperparameters}
    \SetAlgoLined
    \SetNlSty{textbf}{}{:}
    \SetAlgoNlRelativeSize{-1}
    \SetAlCapSkip{0.5em}
    \SetInd{0.2em}{0.5em}
    \SetKwComment{tcp}{//}{}
    \Input{Process model $SN$; Ttrace $T$}
    \Hyperparameters{Number of alignments per subtrace $N$; Window length $L$ %; Window overlap $O
    }
    \Output{Alignment between the trace $T$ and the model $SN$}
    \BlankLine
    Split trace $T$ into subtraces $t_1, t_2, \ldots$, each of length $L$ %with overlap $O$ 
    \; \label{algline:Split}
    Initialize $CurrentMarkings$ with the initial marking of $SN$\;
    Initialize $TopNAlignments$ to hold $N$ empty alignments\;
    Pre-calculate reachable transitions for each marking in the process model\; \label{algline:precalc}
    \For{$i=1$ to $\lceil\frac{\text{length}(T)}{L}\rceil$}{
        Construct synchronous product $SP_i$ from $t_i$ and $SN$\;
        \eIf{$i = \lceil\frac{\text{length}(T)}{L}\rceil$}{
            Initialize $FinalAlignments$ as an empty list\;
            \ForEach{alignment $\alpha$ in $TopNAlignments$}{
                Extend $\alpha$ to include alignment from $RG_{\lceil\frac{\text{length}(T)}{L}\rceil}$ starting from $\alpha$'s final marking\;
                Add this extended alignment to $FinalAlignments$\;
            }
            \textbf{break}\;
        }{
            Initialize $ExtendedAlignments$ as an empty list\;
            Initialize $UniqueFinalMarkings$ as an empty set\;
            \ForEach{alignment $\alpha$ in $TopNAlignments$}{
                \For{$j=1$ to $N$}{
                    Extend $\alpha$ with the alignment from $RG_i$ that minimizes total cost (sum of alignment cost and cost of nonreachable transitions remaining in the trace), leading to a unique final marking\; \label{algline:heuristic_cost}
                    Add the extension to $ExtendedAlignments$\;
                    Add the extension's final marking to $UniqueFinalMarkings$\;
                }
            }
            Sort $ExtendedAlignments$ by total cost and keep top $N$ alignments\;
            Update $TopNAlignments$ with these top $N$ alignments\;
            Update $CurrentMarkings$ with the final markings of these alignments\;
        }
    }
    \Return{The lowest cost alignment from $FinalAlignments$}\;
    \caption{Window Based Conformance Checking}
    \label{alg:WindowBasedConformanceChecking}
\end{algorithm}

\subsection{Sliding Window Mechanism and Iterative Alignment}\label{sec:sliding}

The algorithm starts by decomposing a trace into smaller subtraces. It then initializes two lists: one that records the markings visited during execution, and another for progressively compiling the alignments. Following this initialization, the algorithm performs a preprocessing step for calculating reachable transitions from each model marking  (Lines~\ref{algline:Split}-\ref{algline:precalc}). Each subtrace is then sequntially processed, where a synchronous product between the subtrace and the process model facilitates alignment computations.

For each subtrace, except for the last, the algorithm executes an iterative alignment by exploring the reachability graph $(RG)$ of a subtrace and the process model. This graph facilitates the identification of the most suitable alignments, considering the final marking in the subtrace and any state in the process model as potential endpoints. After each alignment is completed, the algorithm captures the final state it reached, marking the end of that alignment phase. These captured states are then set as the starting points for the alignment of the next subtrace, ensuring a seamless and continuous conformance checking process. As the algorithm progresses, it aggregates the alignments into a list until reaching the last subtrace.

Upon identifying the last subtrace, the algorithm initiates a new list called $FinalAlignments$ for holding the complete alignments. It extends each alignment found in the $TopN Alignments$ list from their most recently recorded state. 

The algorithm sorts the extended alignments based on their total cost, retaining only the top $N$ lowest cost alignments along with their corresponding markings.
The algorithm selects the alignment with the lowest total cost from the $FinalAlignments$ list as the optimal alignment, thereby concluding its execution.

%After extending these alignments to cover the last subtrace, the algorithm finalizes its operations by selecting and returning the alignment with the lowest cost from the $FinalAlignments$ list.

In the next subsection we demonstrate the algorithm's operations over a toy example.

\subsection{Illustrative Example of Algorithm Execution}
\label{sec:illustrativeexample}

\begin{figure}[htbp]
\centering
\includegraphics[width=0.8\columnwidth]{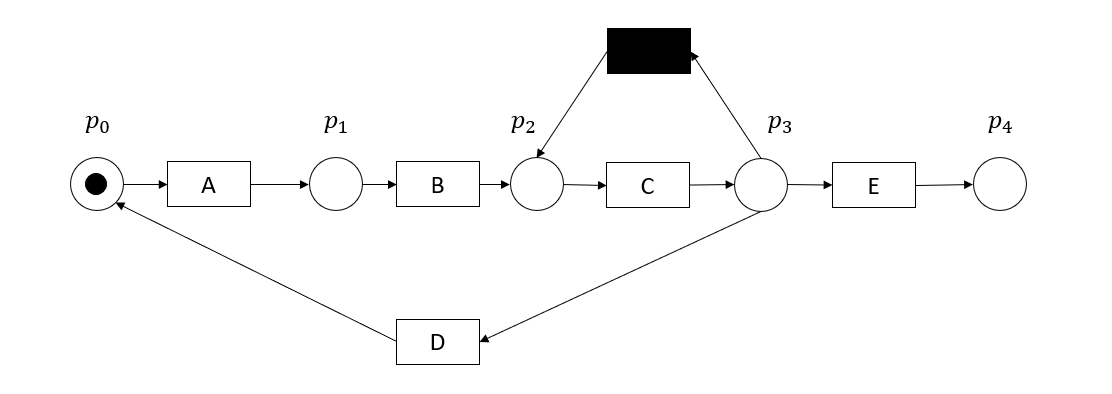}
\caption{An example of a process model.}
\label{fig:processmodelrunningexample}
\end{figure}

\begin{figure}[htbp]
\centering
\includegraphics[width=0.9\columnwidth, height=1cm]{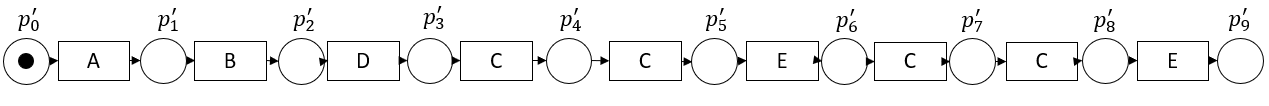}
\caption{An example of a trace model.}
\label{fig:tracemodelrunningexample}
\end{figure}

 Let us manually apply the algorithm over the process model and trace presented in Figures~\ref{fig:processmodelrunningexample}, and~\ref{fig:tracemodelrunningexample}, respectively, to demonstrate how it operates. The model comprises six transitions, including a silent ($\tau$) transition. The trace $\langle ABDCCECCE \rangle$, depicted in the figure as a trace model, includes a sequence of 9 transitions. We set the hyperparameters such that the number of alignments per subtrace is 2, and the window length is 3 so we could demonstrate at least two subtraces.

\begin{figure}[htbp]
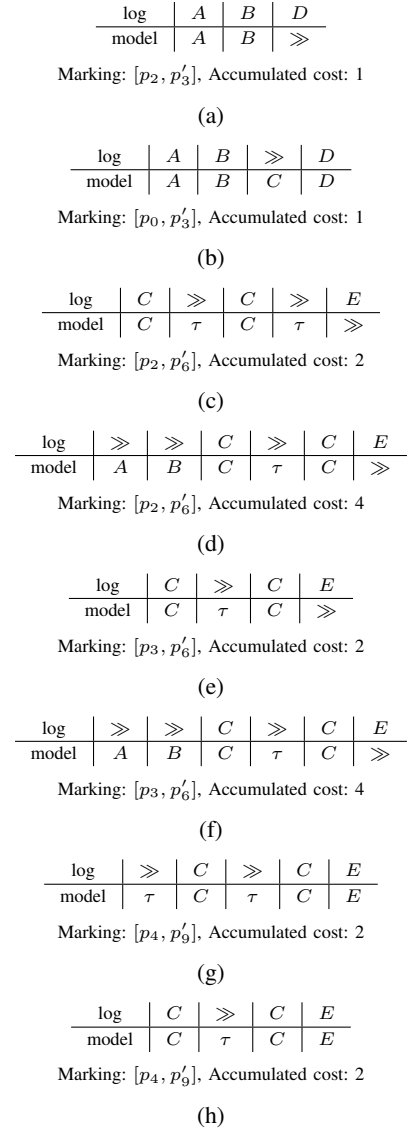

    \centering
    % Table 1
    \begin{subfigure}{\columnwidth}
        \begin{adjustbox}{valign=t}
        \begin{minipage}{\columnwidth}
            \centering
            \scriptsize
            \begin{tabular}{c|c|c|c}
                log & $A$ & $B$ & $D$  \\
                \hline
                model & $A$ & $B$ & $\gg$ 
            \end{tabular}\\[5pt]
            {\scriptsize Marking: $[p_{2}, p'_3]$, Accumulated cost: 1}
        \end{minipage}
        \end{adjustbox}
        \caption{}
        \label{subfig:table1}
    \end{subfigure}

    \vspace{5pt} % Adjust vertical space to separate the groupings visually

    % Table 2
    \begin{subfigure}{\columnwidth}
        \begin{adjustbox}{valign=t}
        \begin{minipage}{\columnwidth}
            \centering
            \scriptsize
            \begin{tabular}{c|c|c|c|c}
                log & $A$ & $B$ & $\gg$ & $D$  \\
                \hline
                model & $A$ & $B$ & $C$ & $D$ 
            \end{tabular}\\[5pt]
            {\scriptsize Marking: $[p_{0}, p'_3]$, Accumulated cost: 1}
        \end{minipage}
        \end{adjustbox}
        \caption{}
        \label{subfig:table2}
    \end{subfigure}

    \vspace{5pt} % Adjust vertical space between the groupings

    % Table 3
    \begin{subfigure}{\columnwidth}
        \begin{adjustbox}{valign=t}
        \begin{minipage}{\columnwidth}
            \centering
            \scriptsize
            \begin{tabular}{c|c|c|c|c|c}
                log & $C$ & $\gg$ & $C$ & $\gg$ & $E$  \\
                \hline
                model & $C$ & $\tau$ & $C$ & $\tau$ & $\gg$ 
            \end{tabular}\\[5pt]
            {\scriptsize Marking: $[p_{2}, p'_{6}]$, Accumulated cost: 2}
        \end{minipage}
        \end{adjustbox}
        \caption{}
        \label{subfig:table3}
    \end{subfigure}

    \vspace{5pt} % Adjust vertical space between the groupings

    % Table 4
    \begin{subfigure}{\columnwidth}
        \begin{adjustbox}{valign=t}
        \begin{minipage}{\columnwidth}
            \centering
            \scriptsize
            \begin{tabular}{c|c|c|c|c|c|c}
                log & $\gg$ & $\gg$ & $C$ & $\gg$ & $C$ & $E$  \\
                \hline
                model & $A$ & $B$ & $C$ & $\tau$ & $C$ & $\gg$ 
            \end{tabular}\\[5pt]
            {\scriptsize Marking: $[p_{2},p'_6]$, Accumulated cost: 4}
        \end{minipage}
        \end{adjustbox}
        \caption{}
        \label{subfig:table4}
    \end{subfigure}

    \vspace{5pt} % Adjust vertical space between the groupings

    % Table 5
    \begin{subfigure}{\columnwidth}
        \begin{adjustbox}{valign=t}
        \begin{minipage}{\columnwidth}
            \centering
            \scriptsize
            \begin{tabular}{c|c|c|c|c}
                log & $C$ & $\gg$ & $C$ & $E$  \\
                \hline
                model & $C$ & $\tau$ & $C$ & $\gg$ 
            \end{tabular}\\[5pt]
            {\scriptsize Marking: $[p_{3},p'_6]$, Accumulated cost: 2}
        \end{minipage}
        \end{adjustbox}
        \caption{}
        \label{subfig:table5}
    \end{subfigure}

    \vspace{5pt} % Adjust vertical space between the groupings

    % Table 6
    \begin{subfigure}{\columnwidth}
        \begin{adjustbox}{valign=t}
        \begin{minipage}{\columnwidth}
            \centering
            \scriptsize
            \begin{tabular}{c|c|c|c|c|c|c}
                log & $\gg$ & $\gg$ & $C$ & $\gg$ & $C$ & $E$  \\
                \hline
                model & $A$ & $B$ & $C$ & $\tau$ & $C$ & $\gg$ 
            \end{tabular}\\[5pt]
            {\scriptsize Marking: $[p_{3},p'_6]$, Accumulated cost: 4}
        \end{minipage}
        \end{adjustbox}
        \caption{}
        \label{subfig:table6}
    \end{subfigure}

    \vspace{5pt} % Adjust vertical space between the groupings

    % Table 7
    \begin{subfigure}{\columnwidth}
        \begin{adjustbox}{valign=t}
        \begin{minipage}{\columnwidth}
            \centering
            \scriptsize
            \begin{tabular}{c|c|c|c|c|c}
                log & $\gg$ & $C$ & $\gg$ &$C$ & $E$ \\
                \hline
                model &$\tau$ & $C$ & $\tau$ & $C$ & $E$
            \end{tabular}\\[5pt]
            {\scriptsize Marking: $[p_{4},p'_9]$, Accumulated cost: 2}
        \end{minipage}
        \end{adjustbox}
        \caption{}
        \label{subfig:table7}
    \end{subfigure}

    \vspace{5pt} % Adjust vertical space between the groupings

    % Table 8
    \begin{subfigure}{\columnwidth}
        \begin{adjustbox}{valign=t}
        \begin{minipage}{\columnwidth}
            \centering
            \scriptsize
            \begin{tabular}{c|c|c|c|c}
                log & $C$ & $\gg$ & $C$ & $E$ \\
                \hline
                model & $C$ & $\tau$ & $C$ & $E$
            \end{tabular}\\[5pt]
            {\scriptsize Marking: $[p_{4},p'_9]$, Accumulated cost: 2}
        \end{minipage}            
        \end{adjustbox}
        \caption{}
        \label{subfig:table8}
    \end{subfigure}

\caption{Markings and alignment costs for subtraces. (\ref{subfig:table1}-\ref{subfig:table2}) show the alignments for the first subtrace. (\ref{subfig:table3}, \ref{subfig:table5}) present alignments starting from $[p_{2},p'_3]$. (\ref{subfig:table4}, \ref{subfig:table6}) present alignments starting from $[p_{0},p'_3]$. (\ref{subfig:table7}-\ref{subfig:table8}) present the lowest cost alignments for the last subtrace starting from $[p_{3},p'_6]$ and $[p_{2},p'_6]$, respectively.}
\end{figure}

As described in Algorithm~\ref{alg:WindowBasedConformanceChecking}, the trace is split into the 3 subtraces $\langle ABD \rangle$, $\langle CCE \rangle$, and $\langle CCE \rangle$. It then prepares arrays for storing intermediate results and computes the reachable transitions from each model marking. For example, from the initial marking $[p_{0}]$, the set of reachable transitions is $\{A, B, C, D, E\}$. The algorithm computes the two best alignments for the first subtrace $\langle ABD \rangle$. The initial marking is $[p_0,p'_0]$, and the final markings are defined as any model marking as long as the subtrace's token reaches its final place, that is $[\cdot,p'_3]$ for the first subtrace. The two best alignments are presented in Figures~\ref{subfig:table1} and~\ref{subfig:table2}, correspond to markings $[p_{2},p'_3]$ and $[p_{0},p'_3]$, respectively. Each alignment incurs a cost of 1 due to a required nonsynchronous move. The remaining trace transitions that have $C$ and $E$ labels are reachable from model markings $[p_2]$ and $[p_0]$, thus the lower bound on the marginal cost is 0. %The cost and alignment of the final transition in the subtrace are omitted (illustrated by a red vertical line). Additionally, adhering to the hyperparameter setting, 
Both alignments and their resultant final markings are recorded for subsequent processing.

Next, the algorithm aligns the second subtrace $\langle CCE \rangle$, starting from the two final markings of the first trace, that is once from marking $[p_{2}, p'_{3}]$ and once from $[p_{0}, p'_{3}]$. This results in two unique optimal alignments from each starting point, as shown in Figures~\ref{subfig:table3} and ~\ref{subfig:table5} (from $[p_{2}, p'_{3}$]) and Figures~\ref{subfig:table4} and ~\ref{subfig:table6} (from $[p_{0},p'_3]$). The best two alignments starting from $[p_{2}, p'_{3}$]) each incur a cost of 1, leading to a total cost of 2 (taking into account the cost of the first subtrace). Interestingly, there is a perfect local alignment that costs 0 locally: perform synchronous move with $C$, then a silent model move, again a synchronous move with $C$ and then a synchronous move with $E$ with a final marking of $[p_{4},p'_6]$. This alignment is disqualified by the algorithm since the lower bound on the marginal cost is 3 since there are three mandatory trace transitions in the final subtrace that are not reachable from model marking $[p_4]$. 

Alignments from $[p_{0},p'3]$ accumulate a total cost of 4. The algorithm then retains the two lowest-cost aggregated alignments, which have distinct final markings of $[p{2},p'6]$ and $[p{3},p'6]$, both originating from the earlier marking $[p{2},p'_3]$. As noted, none of these alignments included activity $E$ as a synchronous move due to the marginal cost. Firing activity $E$ in the process model would result in realizing the remaining trace transitions as nonsynchronous moves, which would incur additional marginal costs.
 
In the last subtrace, the algorithm aligns $\langle CCE \rangle$, starting from markings $[p_{2},p'_6]$ and $[p_{3},p'_6]$. As this is the last subtrace, the search focuses on a single optimal alignment for each starting marking, reflecting the final integrated marking of both the trace and model. Both alignments accumulate a total cost of 2, as can be seen in Figures~\ref{subfig:table7} and~\ref{subfig:table8}. Both alignments have the same cost, so one of them is chosen arbitrarily (in this example, we selected the alignment from marking $[p_{2},p'_6]$). The algorithm's final output is the combined sequence of alignments leading, in this case, to the optimal alignment with a conformance cost 2. %from Figures~\ref{subfig:table1},~\ref{subfig:table3}, and~\ref{subfig:table8}, achieving an overall cost of 2.

\section{Complexity Analysis}\label{sec:complexity}

During recent years, several alignment-based conformance checking algorithms have been proposed. They were coded in different programming languages, by different programmers and tested on different machines with a variety of datasets and with respect to different process models making it difficult to compare them in terms of computational times. Moreover, there are almost no datasets that contain long traces with hundreds or thousands of events. We, therefore, offer insights into the complexity of the suggested algorithm to demonstrate its scalability for handling very long traces compared to other alternatives.

Most alignment-based conformance checking methods employ search techniques such as $A^*$ to find the optimal alignment. The complexity of finding such an optimal alignment is exponential in the worst case, with the base of the exponent being the branching factor of the search algorithm, denoted as $b$ and the exponent is $d$, the depth of the solution, which is at least as long as the length of the trace. Accordingly, the worst-case time complexity is $O(b^d)$, which is enormous when $d$ is large.

Recent optimal alignment methods include improvements for accelerating their time performance but do not fundamentally change their worst-case complexity. For example, \cite{van2018efficiently} introduces a method that uses the extended marking equation to reduce the search space. This approach leverages structural information of the Petri net to prune the state space, yet the overall complexity remains exponential in the length of the solution.
Similarly, the authors of \cite{casas2024reach} propose %an algorithm named REACH, which uses a preprocessing step to compute mandatory transitions required to reach the final marking. 
a heuristic that do not require solving the marking equation for each state and thus reduces the computational burden. Despite the heuristic, the complexity grows exponentially with the solution's length since the number of searched alignments can grow exponentially.

Consequentially, these promising algorithms would not be able to handle very long traces in reasonable times. 
The approach suggested in this paper mitigates the exponential growth by decomposing the trace into fixed-length subtraces of length $L$ and by aligning each subtrace separately. The number of subtraces, $W$, is determined by $W = \lceil N / L \rceil$ where $N$ is the length of the complete trace. %where $O$ is the  overlap between consecutive subtraces. 
For long traces, especially those generated by sensors which motivate this work, the shortest path through the process model, $L_m$, is much smaller than the trace length, that is $N \gg L_m$.

Here are the principles guiding the complexity analysis:

\begin{enumerate}
    \item Trace partitioning: The trace is divided into $W$ subtraces, each of length $L$ (apart from the last one). %The overlap $O$ between consecutive subtraces can be adjusted to balance computational efficiency and alignment accuracy.
    \item Alignment of subtraces: Each subtrace is aligned independently with the process model. The depth of the solution for each subtrace would be approximately $L$ and not more than $2L$, thus the complexity of aligning a subtrace is $O(b^{2L})$.
    \item Combining alignments: The alignments of the subtraces are combined iteratively. Since the number of subtraces $W$ is proportional to $N / L $, the overall complexity increases linearly with the number of subtraces.
\end{enumerate}

It is important to note that the complexity of aligning a single subtrace remains exponential but we limit the exponent size by selecting $L$ such that subtraces would align very quickly. The alignment of each subtrace, except from the last one, completes when the token reaches the last place in the subtrace model without constraining the model's marking. In other words, we do not require tokens to reach the final place in the process model. Under such a situation the solution depth will not exceed $2L$ since otherwise it would be favorable to perform $L$ log moves and complete the alignment. Thus, the complexity would be $O(b^{2L})$, where $L$ is selected by the user to facilitate fast subtrace computational times. Therefore, the total complexity of our approach is the complexity per intermediate subtrace multiplied by $W-1$. The complexity of the last subtrace is bounded by an exponent of $\max(2L,L+L_m)$, Therefore:

\[ O\left(\frac{N-L}{L} \cdot b^{2L} +  b^{\max(2L,L+L_m)}\right) \]

That is, the complexity is linear with the number of subtraces, making the sliding window approach significantly more scalable for long traces compared to traditional alignment-based methods. The solution times from the experimental study support this analysis.

\section{Empirical Evaluation}
\label{sec:empirical_evaluation}

%This section presents empirical evidence to demonstrate the effectiveness and efficiency of our proposed approach. Through a series of experiments encompassing diverse scenarios, we rigorously assess our algorithm's performance.

We performed a series of experiments to evaluate the scalability and performance of the suggested method over long traces. %and its performance in terms of deviation from the optimal conformance costs. 
Bear in mind that the suggested approach is tailored for long traces with hundreds and thousands of events which are typically generated by sensors and prediction models. Unfortunately, classical process mining datasets typically do not contain long traces so they cannot fully showcase the scalability of the suggested method. This dictated an experiment design in which the first set of experiments was used to evaluate the approach compared to an optimal method.  
The second set of experiments tested the approach on publicly available computer vision datasets that generate traces with thousands of events. None of the optimal conformance checking approaches could handle these traces, but we demonstrated that the suggested approach can perform conformance checking within reasonable computational times, as indicated by the complexity analysis.

The algorithms were coded in Python and the experiments were run on a machine with Intel Xenon processor E5-2650 @2.20GHz with 24 cores (each with 2 threads). To control the evaluation environment the experiments were run within docker containers configured with 10 CPUs and 30 GB of allocated memory.

%to scales linearly with trace length, contrasting with traditional approaches that often exhibit exponential scaling when processing entire traces. We measure scalability by tracking the number of nodes explored during the alignment search.
%\item \textbf{Near-Optimal Performance:} Even when dividing the trace into small segments, our algorithm consistently delivers results close to the optimal solution across a wide range of datasets and models. We assess performance by comparing the cost of our alignments to the optimal alignments. While we record execution times for completeness, absolute timings are not a primary focus, as they can be influenced by implementation details, hardware, and software optimizations.
%\end{enumerate}

\subsection{Classic Datasets}

Following the methodology used in previous works (e.g., \cite{van2018efficiently}), we focus on demonstrating the quality of our results while using a much smaller search space than traditional methods. We do not present the datasets noted by `clean' in which the model and the log are completely fitting since these datasets for which our approach performed perfectly do not pose a challenge.

We filtered the datasets, to include their longest traces (i.e., those longer than 100, when existing, or otherwise longer than 80 transitions). To simulate settings in which the trace length is much longer than that of the subtrace, we intentionally used small windows, keeping them several times shorter than the traces (i.e., \(5 \leq L \leq 50\)). This is far from ideal since such small windows introduce significant overhead due to the repeated construction and exploration of the synchronous products which can hurt the performances. This overhead can be significantly reduced by selecting a longer window, incremental synchronous product computation and caching mechanisms, which are beyond the scope of this work. When a dataset did not include a process model, we discovered one. For this, we utilized approximately 10\% of the traces from the dataset. Table \ref{tab:classic_datasets} summarizes our findings. The results in the table are presented in reference to $A^*$.

\subsection{Long Traces of Food Preparation Datasets}

For the second part of our experiments, we evaluate our algorithmic approach on publicly available food preparation datasets (see \cite{bogdanov2023sktr}). We use three datasets: 50 Salads, GTEA, and Breakfast. Traces of these datasets reach thousands of events, which cannot be handled by optimal alignment approaches. %We discovered a model based on part of the data to simulate a situation in which the model would not always fit the data. 
For these datasets, Table \ref{tab:video_datasets} reports the average number of explored states and the associated conformance cost. For these extensive traces, we allocated a time limit of 120 sec per trace and all were solved within this timeframe. For the Breakfast dataset, which contains many unique and highly dissimilar traces, the conformance cost for a significant portion of the traces was very high, as expected, see also in Table \ref{tab:food}.
  
\begin{table}[ht]
\centering
\caption{Results for classic datasets. The columns from left to right describe the dataset characteristics and the average results relative to the results achieved by using $A^*$. The rightmost columns show the explored nodes, the conformance cost difference ($\Delta$ cost), and the CPU usage.}

\label{tab:classic_datasets}
\resizebox{\columnwidth}{!}{%
\begin{tabular}{|c|c|c|c|c|c|c|}
\hline
Dataset & Cases (\#) & Tr. length & Optimal (\%) & Explored nodes (\%)  & $\Delta$ cost (\%) & CPU (\%)  \\
%\hline
%pr\_1912\_l2\_clean & 4 & 63 & 100 & 100 & 63 & 0 & 0 & 0.04 & 0.03\\ 
\hline
% pr\_1912\_l2\_noise & 4 & 61 & 100 & 2.8 & 0 & 8.3 \\ 
% \hline
%pr\_1912\_l3\_clean & 444 & 69 & 100 & 100 & 70 & 0 & 0 & 0.04 & 0.03\\ 
%\hline
% pr\_1912\_l3\_noise & 459 & 70.1 & 46.6 & 2.7 & 9.3 & 5.6 \\ 
% \hline
%pr\_1912\_l4\_clean & 824 & 82.7 & 100 & 82 & 82 & 0 & 0 & 0.06 & 0.05\\ 
%\hline
pr\_1912\_l4\_noise & 95 & 107 & 95.8 & 18.6 & 0.6 & 11 \\ 
\hline
% pr\_1908\_l3\_noise & 10 & 61.9 & 100 & 6.3 & 0 & 15.8  \\ 
% \hline
pr\_1908\_l4\_noise & 2 & 84 & 100 & 24 & 0 & 16 \\ 
\hline
% pr\_1151\_l2\_noise & 3 & 65 & 100 & 13.6 & 0 & 28.6 \\ 
% \hline
pr\_1151\_l3\_noise & 92 & 87 & 95.6 & 17 & 1.0 & 19 \\ 
\hline
%pr\_1151\_l4\_clean & 747 & 86 & 100 &86 & 86 & 0 & 0 & 0.06 & 0.05\\ 
%\hline
pr\_1151\_l4\_noise & 422 & 99 & 90 & 15.2 & 1.0 & 14 \\ 
\hline
%pr\_1244\_l2\_clean & 5 & 61.4 & 100 &61 & 61 & 0 & 0 & 0.07 & 0.04\\ 
%\hline
% pr\_1244\_l2\_noise & 12 & 61.5 & 100 &7.2 & 0 & 26.9 \\ 
% \hline
pr\_1244\_l3\_noise & 2 & 102 & 100 & 30 & 0 & 22 \\ 
\hline
pr\_1244\_l4\_noise & 226 & 111 & 87.1 & 13.2 & 2.2 & 8 \\ 
\hline
prEm6 & 535 & 106 & 100 & 14.8 & 0 & 12    \\ 
\hline
Sepsis & 5 & 129 & 100 & 7.2 & 0 & 2.7 \\ 
\hline
BPIC\_2012 & 825 & 190 & 99.8 & 50 & 0.9 & 15.6    \\ 
\hline
BPIC\_2017 & 204 & 114 & 99.0 & 27.5 & 0.9 & 13.9   \\
\hline
% BPIC\_2019 &  &  &  &  &  &     \\ 
\end{tabular}
}
\end{table}

\begin{table}[ht]
\centering
\caption{Results for food preparation datasets. The two left-hand columns describe the dataset characteristics and the other columns present the average results.}
\label{tab:food}
\resizebox{\columnwidth}{!}{%
\begin{tabular}{|c|c|c|c|c|c|c|c|}
\hline
Dataset &  Cases (\#) & Tr. length & Explored nodes (\#) & Cost & CPU (sec)\\
\hline
% Your data here; example row below
GTEA
 & 28 & 1301 & 29674 & 0.1 & 3.3 \\
 \hline
Breakfsast
 & 1008 & 2005 & 99065 & 371.4 & 19.0 \\
 \hline
50 Salads
 &  40 & 5945 & 591681 & 1.7 & 98 \\
 \hline

\end{tabular}
}
\end{table} \label{tab:video_datasets}

\section{Related Work}
\label{sec:literature_review}

Improving the performance of conformance checking algorithms is an active research area for which we review selected publications. 

Van Dongen~\cite{van2018efficiently} developed a conformance checking approach that uses the extended marking equation to expedite the search for optimal alignments. The approach was found favorable compared to other alternatives in terms of computational times. Still, it faces exponential growth in complexity as trace length increases, rendering it impractical for extremely long traces.

Similarly, \cite{casas2024reach} proposed a new algorithm named REACH for computing optimal alignments. In contrast to previous methods, the heuristic used by REACH uses a preprocessing step in which the algorithm computes mandatory transitions within the process model. This heuristic enables avoiding from solving the marking equation, significantly reducing the computation overhead. Despite its improvement in computation times, REACH also struggles with the exponential growth of the search space with increasing trace lengths and cannot handle very long traces.

Hierarchical conformance checking methods \cite{munoz2013hierarchical} facilitate effective identification and quantification of discrepancies between observed and modeled behaviors. This approach is particularly beneficial for intricate event logs, as it constrains the size of the conformance instances, thereby reducing the complexity of the analysis. For the specific aim of estimating fitness and precision, recent techniques \cite{wang2022novel, burattin2019fifty} have introduced new perspectives to process mining by improving the differentiation of traces, which is especially beneficial for handling long traces.

In another line of works, \cite{lee2018recomposing, lee2017replay} introduced a replay using recomposition technique, applying decomposition \cite{van2013decomposing} to compute the fitness value efficiently. Users can configure the balance between accuracy and computation time to get a fitness interval under time constraints. This may be useful for fast computations but can limit insights since the technique does not compute alignments explicitly. While our approach shares common ideas, it focuses on trace decomposition rather than model decomposition. Moreover, at each subtrace alignment, the model markings that correspond to the $k$ best alignments are reserved to facilitate smart recomposition of the complete alignment. When structures leading to alignment errors are identified, the number of best alignments and the subtrace lengths can be tuned.

Other studies, \cite{bauer2022sampling,fani2020conformance}, use selection and sampling strategies for approximating the conformance of a log while providing statistical guarantees on the conformance level. This approach, which is appropriate when seeking to estimate conformance within an organization, cannot fit the task of conformance checking of individual traces.
%The token-based replay technique \cite{berti2021novel} offers improvements in scalability and diagnostic accuracy for the analysis of complex models and large logs. Although this method demonstrated enhanced performance, it does not consistently exceed alignment-based alternatives and lacks certain assurances, such as fitness.

% Advances in computational techniques and the integration of machine learning into conformance checking have yielded efficient conformance checking techniques. \citeauthor{valencia2021empowering}~\cite{valencia2021empowering} propose process mining architectures that utilize Big Data infrastructure. Specifically, they show how different types of alignments can be applied in a distributed paradigm, leading to better utilization of computational resources. \citeauthor{peeperkorn2020conformance}~\cite{peeperkorn2020conformance} use neural network-based embeddings, and \citeauthor{bloemen2018symbolically}~\cite{bloemen2018symbolically} and \citeauthor{reissner2020efficient}~\cite{reissner2020efficient} explore symbolic and heuristic strategies, yielding improvements in efficiency and reliability. Moreover, real-time process mining \cite{guo2021online, stertz2020temporal} provides instant feedback and effectively manages real-time data, essential for continuous process monitoring and adaptation in dynamic business environments.

Despite the progress in conformance checking, efficiently processing long traces remains a challenge, and current techniques fall short in handling long traces effectively. This work proposes a promising sliding window approach that can handle very long traces in realistic computational times while achieving near-optimal results.

\section{Conclusion and Future Directions}\label{sec:conclusion}
This paper presents a new conformance checking approach that is specifically designed to handle long traces that are generated by sensors and prediction models. Conformance checking of such traces has been identified as one of the biggest process mining challenges \cite{beerepoot2023biggest}. The suggested approach satisfies the scalability challenge by decomposing traces into manageable subtraces and iteratively aligning them with the process model. In that, the search space is reduced while maintaining the interpretability of alignment-based methods. 
By incorporating global information that captures the structural properties of both the trace and the model, the approach guides intelligent alignment decisions, further enhancing accuracy.

We conducted a complexity analysis and experimental evaluations to showcase the algorithm's scalability. The theoretical complexity analysis shows a linear growth of the search space with the number of windows and the trace length, a significant improvement over the exponential complexity of traditional methods. 
We used multiple datasets to demonstrate the near-optimal performance of the suggested approach that finds the optimal alignment in over $96\%$ of the cases. We validated the approach for traces with thousands of transitions using publicly available computer vision food preparation datasets. For such datasets, which existing conformance checking methods cannot handle, the proposed approach achieved solutions for all the cases within a reasonable time limit of 2 minutes.

Future research directions include: 1) exploring how to adjust window sizes dynamically based on the characteristics of the trace, 2) incorporating domain-specific heuristics to make informed alignment decisions, to improve the overall accuracy, and to 3) extend the approach to handle multi-dimensional event data and concurrent activities within traces. %Investigating the integration of our method with other conformance checking techniques, such as token-based replay or hierarchical approaches, could also lead to more powerful and versatile solutions for large-scale process mining applications.

% *** a paragraph on alignment based conformance checking with definitions of a trace model, synchronous product, cost function and optimal alignment (Definitions 12-16 in the book of Mathias about conformance checking can do - maybe shorten or combine)

% *** a paragraph on the principles of A-star.

% *** then, a definition of decomposed trace

% ***maybe a definition of a decomposed synchronous product

% *** a motivating example

%nlike the standard approach, where computational complexity increases exponentially with the length of the trace \cite{van2022process}, our approach, by dividing the trace into fixed-length subtraces, ensures that complexity scales linearly with the number of subtraces into which the original trace was segmented. Figure \ref{fig:computationalcomplexity} demonstrates the comparative computational complexities of our method versus the traditional alignment-based approach.
%\printbibliography
\bibliographystyle{IEEEtran}
\bibliography{main}

\end{document}